\documentclass[conference]{IEEEtran}

\IEEEoverridecommandlockouts
\usepackage{cite}
\usepackage{amsmath,amssymb,amsfonts}
\usepackage{algorithmic}
\usepackage{graphicx}
\usepackage{textcomp}
\usepackage{xcolor}
\usepackage{float}

\usepackage{color}
\usepackage{amsthm}
\usepackage{subcaption}
\usepackage{multicol}
\usepackage{enumitem}
\usepackage[normalem]{ulem}

\def\l{\lambda}

\def\RR{\mathbb{R}}

\def\ZZ{\mathbb{Z}}

\def\cN{\mathcal{N}}

\def\cP{\mathcal{P}}

\def\BA{\boldsymbol{A}}
\def\BD{\boldsymbol{D}}

\def\BL{\boldsymbol{L}}

\def\BS{\boldsymbol{S}}

\def\BU{\boldsymbol{U}}
\def\BW{\boldsymbol{W}}
\def\BX{\boldsymbol{X}}

\def\Bf{\boldsymbol{f}}
\def\Bu{\boldsymbol{u}}

\def\Bz{\boldsymbol{z}}

\def\BPhi{\boldsymbol{\Phi}}

\def\Bmu{\boldsymbol{\mu}}
\def\Bsigma{\boldsymbol{\sigma}}
\def\BSigma{\boldsymbol{\Sigma}}
\def\Bphi{\boldsymbol{\phi}}
\def\Bpsi{\boldsymbol{\psi}}

\newcommand{\abs}[1]{\left| #1 \right|}

\newcommand{\norm}[1]{\left\lVert#1\right\rVert}

\def\BibTeX{{\rm B\kern-.05em{\sc i\kern-.025em b}\kern-.08em
    T\kern-.1667em\lower.7ex\hbox{E}\kern-.125emX}}
\begin{document}

\title{Encoding robust representation for graph generation\\
\thanks{This research has been supported by NSF award DMS-18-30418.}
}

\author{\IEEEauthorblockN{Dongmian Zou}
\IEEEauthorblockA{\textit{Institute for Mathematics and its Applications} \\
\textit{University of Minnesota, Twin Cities}\\
Minneapolis, USA \\
dzou@umn.edu}
\and
\IEEEauthorblockN{Gilad Lerman}
\IEEEauthorblockA{\textit{School of Mathematics} \\
\textit{University of Minnesota, Twin Cities}\\
Minneapolis, USA \\
lerman@umn.edu}
}

\maketitle

\begin{abstract}
Generative networks have made it possible to generate meaningful signals such as images and texts from simple noise. Recently, generative methods based on GAN and VAE were developed for graphs and graph signals. However, the mathematical properties of these methods are unclear, and training good generative models is difficult. This work proposes a graph generation model that uses a recent adaptation of Mallat's scattering transform to graphs. The proposed model is naturally composed of an encoder and a decoder. The encoder is a Gaussianized graph scattering transform, which is robust to signal and graph manipulation. The decoder is a simple fully connected network that is adapted to specific tasks, such as link prediction, signal generation on graphs and full graph and signal generation. The training of our proposed system is efficient since it is only applied to the decoder and the hardware requirements are moderate. Numerical results demonstrate state-of-the-art performance of the proposed system for both link prediction and graph and signal generation.
\end{abstract}


\section{Introduction}\label{sec:intro}
Generative neural networks have 
been successfully applied to various tasks such as the generation of images and texts.
Their development is based on fruitful methods of deep learning, such as convolutional and recurrent neural networks. These and other methods of deep learning, which were initially developed for problems in the Euclidean domain\footnote{We remark that the Euclidean and graph domains include scenarios whose underlying datasets have Euclidean and graph structures, respectively.}, have been successfully generalized to address supervised learning tasks in the graph domain. 
In particular, a variety of graph convolutional networks have been developed, including networks with prescribed parameters \cite{zou2018graph, gama2018diffusion} and trained networks \cite{henaff2015deep, defferrard2016convolutional, kipf2016semi, chen2017supervised}. It is natural to use these tools to build generative models in the graph domain.

Most generative graph networks directly use standard graph networks by following either of the following two generative frameworks: the generative adversarial network (GAN) \cite{goodfellow2014generative} and the variational auto-encoder (VAE) \cite{kingma2013auto}. 
In a GAN, a generator and an auxiliary adversarial discriminator are trained together. On the other hand, in VAE, an encoder and a decoder (or generator) are both trained according to Bayesian models. Both frameworks contain two components (generator and discriminator or encoder and decoder), where each of them requires training. For GAN, training two components corresponds to a difficult min-max problem. On the other hand, training the two components in VAE can be described as a relatively easier non-convex minimization problem. However, it is a crude approximation to the motivating variational inference formulation.
Given an encoding process with guaranteed mathematical properties, one can focus on training only the decoder, which in this case is the generator.
In the Euclidean domain, \cite{angles2018generative} uses the scattering transform as an encoder, which is robust to deformations of input signals, and learn a generative model by minimizing the $l_1$-loss for reconstructing the training images. We adopt a similar method, using a graph scattering transform as an encoder for graph signals, which is robust to signal and graph manipulations, and train neural networks corresponding to respective tasks. Note that the graph scattering transform can be either carried out in the spectral domain \cite{zou2018graph} or in the graph domain \cite{gama2018diffusion}. More specifically, spectral-domain wavelets \cite{hammond2011wavelets} are one-dimensional wavelets applied to the eigenvalues of the graph Laplacian, whereas graph-domain   wavelets, or diffusion wavelets \cite{coifman2006diffusion}, are multi-scale functions on the vertices of the graph that use dyadic powers of the diffusion operator. The advantage of the former graph scattering transform is its guaranteed robustness to signal manipulation, which is a consequence of its energy preservation \cite{zou2018graph}. We therefore emphasize this transform here, but we also experiment with the other transform. We also remark that the spectral-domain wavelets of the former transform allow flexible choices of wavelet functions. 
We note that a conventional generative scattering network \cite{angles2018generative} is not as competitive as state-of-the-art results based on GAN and VAE for image generation tasks. This is probably due to the complexity of some Euclidean-type datasets and their nontrivial high frequency components. Nevertheless, graph-type datasets have a discrete nature, and they often do not exhibit high frequency components. Therefore, the graph scattering transform might be competitive and efficient for specific generative tasks in the graph domain.

We consider three types of graph generation tasks:
\begin{itemize}[leftmargin=*]
    \item \emph{Link prediction}: In this task, one is interested in predicting whether two vertices from the same graph are connected. The common input includes a graph with missing edges and features of vertices. The goal is to decide whether an edge exists between any pair of vertices. This can be viewed as generating a graph from a latent representation of the partially available graph.  

    A well-known application of this task is the prediction of citations. Common citation datasets were collected by \cite{sen2008collective} and further pre-processed by \cite{kipf2016variational}. These datasets of publications and citations contain features for each publication as well as citation linkage, which are modeled by an undirected graph. In the pre-processed data, one only partially knows the citation linkage and the task is to recover the citations for all pairs of publications.
    
    \item \emph{Signal generation on graphs}: In this task, the graph is fixed and the set of vertices and edges is known. The goal is to generate signals on the given graph. Although we are unaware of convincing data of this type, we believe that this is a potentially useful task. Among the three tasks we review here, it is most similar to generation tasks in Euclidean domains. We can thus enforce some graph structure in special Euclidean-type or grid-type datasets.  Here we pursue this idea with the Fashion-MNIST dataset of images of clothing items \cite{xiao2017fashion}. We do not consider the domain of a $28 \times 28$ pixel image as Euclidean, but associate to it a graph, where each pixel is connected by an edge with its nearby neighbors. Each image is then a signal on this graph. One then needs graph-based methods for generating these signals. 
    
    \item \emph{Graph and signal generation}: In this task, one needs to generate both the graph structure and the signals on the graph. An interesting application is the design of chemical molecules. A network learns from a given dataset, such as QM9 \cite{ramakrishnan2014quantum}, to generate both the atoms (signals on vertices) and the bonds (edges) from a latent sample. This can be used as a purely machine learning-based approach for the design of new drugs \cite{olivecrona2017molecular}.
    
\end{itemize}

Our proposed method is easy to implement. Furthermore, the adjustment of the structure of the decoder to the three types of tasks does not require a lot of effort. 
Unlike GAN or VAE, the model in this paper only requires  training the generator. Meanwhile, there is flexibility in designing the graph wavelets and in choosing the decoder structure. We believe it is a highly adaptable method for various graph tasks and indeed our numerical experiments demonstrate competitive results.

\section{Background}\label{sec:background}

We overview previous relevant works as follows: \S\ref{sec:mallat} reviews generative scattering networks, 
\S\ref{sec:GCN} reviews graph convolutional networks, and \S\ref{sec:graph_gen_net} reviews some recent graph generative models.

\subsection{Scattering networks for generative models} \label{sec:mallat}

The generative scattering network \cite{angles2018generative} can be considered as an encoder-decoder system in  which one only needs to train the decoder. The feature extraction part of the encoder is a scattering transform \cite{mallat2012group, bruna2013invariant} with fixed parameters. It provides multi-scale signal representation, which is Lipschitz continuous with respect to small deformations. The next part of the encoder aims to map the transformed signals into samples of a Gaussian latent variable. We refer to this step as Gaussianization. We later describe in \S\ref{sec:scattering} two Gaussianization methods. 

\begin{figure}[t]
    \centering
    \includegraphics[width=.8\linewidth]{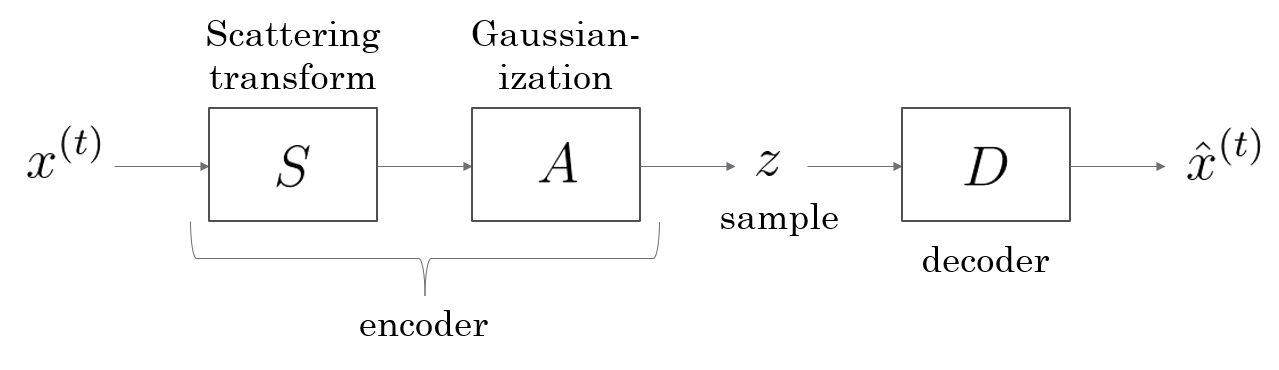}
    \caption{The structure of a generative scattering network.}
    \label{fig:mallat}
\end{figure}

The decoder $D$ can be taken to be a multi-layer perceptron (MLP) and is trained by minimizing the reconstruction loss.
Fig.~\ref{fig:mallat} shows the structure of a generative scattering model. In order to generate samples, initial samples are generated from the latent Gaussian variable and then the decoder is applied to them resulting in the final samples. 

\subsection{Graph convolutional networks}
\label{sec:GCN}
Convolution is a key contributor for the recent success of deep learning. In the Euclidean domain, convolutional networks are helpful in learning multi-scale representations. The same idea was introduced to graphs by exploiting the spectral graph representation, that is, the spectral decomposition of the graph weight matrix or graph Laplacian \cite{bruna2013spectral, henaff2015deep}.  A common proposal for a graph convolution uses pointwise multiplication of the graph Fourier-transformed signals, where the graph Fourier transform uses the basis of the spectral decomposition of the Graph Laplacian in place of the discrete Fourier basis. One can apply nonlinear functions such as the ReLU for each graph vertex.
A variety of good approximations to the spectral approach are able to speed up the spectral decomposition process and maintain accuracy \cite{defferrard2016convolutional, kipf2016semi}.

A special type of graph convolutional network (GCN) is the graph scattering network \cite{zou2018graph}, which does not require training and was proved to be approximately invariant to permutations and stable to sufficiently small signal or graph manipulations. A graph scattering network uses graph wavelets \cite{hammond2011wavelets} defined on the eigenspace of the graph Laplacian to construct multi-layer models. Alternatively, \cite{gama2018diffusion} construct graph scattering transforms using an earlier graph wavelet transform \cite{coifman2006diffusion}. In general, the graph wavelets of \cite{hammond2011wavelets} are more flexible as one can choose different kinds of wavelets on the spectral domain according to different tasks. While the graph wavelets of \cite{coifman2006diffusion} are not flexible, they use the normalized graph Laplacian and the corresponding diffusion map and metric, which might be natural for particular applications.

\subsection{Graph generative networks}
\label{sec:graph_gen_net}
Several recent papers address graph generation networks with either a GAN or a VAE structure, where specific designs are often needed for different applications. 

Recent graph generative models with a GAN structure include NetGAN \cite{bojchevski2018netgan}, GraphGAN \cite{wang2017graphgan} and MolGAN \cite{de2018molgan}. NetGAN aims to generate graphs with properties, such as max degree and triangle count, that are similar to training samples. GraphGAN generates distribution of edges in order to solve node classification and recommendation problems whose tasks are very similar to that of the link prediction problem. MolGAN is designed for molecule generation and generates signals with respect to both vertices and edges. 

Recent graph generative models with a VAE structure include VGAE \cite{kipf2016variational}, GraphVAE \cite{simonovsky2018graphvae} and JT-VAE \cite{jin2018junction}. VGAE aims to solve the link prediction problem by completing the adjacency matrix. GraphVAE is designed for molecule generation. It generates the adjacency matrix as well as the vertex and edge features. JT-VAE is specifically designed for generating complex molecules while enforcing validity.

\section{Graph generative scattering network}\label{sec:scattering}

Existing graph generation networks require training either GAN or VAE, which is a difficult task. Furthermore, the design of a graph generation network is often complex \cite{bojchevski2018netgan, de2018molgan, jin2018junction} and its hyperparameter selection might be difficult. In view of these obstacles, we propose here the graph generative scattering network. It is composed of two components: an encoder and a decoder. The encoder is a graph scattering network \cite{zou2018graph}, which is followed by a Gaussianization step. It produces a latent Gaussianized representation for the graph signal. It is also used to form a latent {``}Gaussian distribution.{''} The parameters of the graph scattering network are predetermined, unlike the parameters of a generative auto-encoder which are learned. 
The decoder is trained by using the Gaussianized latent representation of the data and minimizing a loss function chosen according to the corresponding task. In all of the graph-specific tasks we mentioned in \S\ref{sec:intro}, the decoder can be taken to be a network with fully-connected layers, whose structure is determined by the specific task. Generation is obtained by applying the trained decoder to the latent Gaussian distribution. More details on forming the encoder and decoder are provided in \S\ref{sec:graph_scat_encode} and \S\ref{sec:graph_scat_decode}, respectively. 

\subsection{Details of the Encoder}
\label{sec:graph_scat_encode}
In order to fully understand the formation of the encoder, we review the graph scattering network of \cite{zou2018graph} and explain how to form a Gaussian distribution from its output. We consider a graph $G = (V, E)$ with $\abs{V} = N$ vertices. A signal in $L^2(V; \RR^K)$ can be regarded as a matrix $\BX \in \RR^{N \times K}$. The scattering transform can be regarded as a function that acts on the columns of $\BX$.
Let $\BL \in \RR^{N \times N}$ be the unnormalized graph Laplacian $\BL = \BD - \BW$, where $\BD$ is the diagonal matrix of degrees and $\BW$ is the weight (adjacency) matrix whose $(n,m)$-th entry is the weight of the edge connecting vertices $v_n$ and $v_m$. Its spectral decomposition can be written as 
$    \BL = \sum_{l=0}^{N-1} \l_l \Bu_l \Bu_l^*$, with $0 = \l_0 \leq \cdots \leq \l_{N-1}$. 
We assume a limiting scale, $J \in \ZZ$, and dyadic wavelets, $\phi$ and $\psi$, satisfying $|\hat{\phi}_{-J}|^2 + \sum_{j > -J} |\hat{\psi}_j|^2 = 1$, 
where $$\hat{\psi}_j (\omega) = \hat{\psi}(2^{-j} \omega) \ \text{ for } \ j > -J \ \text{ and } \ \hat{\phi}_{-J} (\omega) = \hat{\phi}(2^{-J} \omega).$$
For $\Bf \in \RR^N$, the graph wavelet transform \cite{hammond2011wavelets}
is
\begin{equation*}
\begin{aligned}
\Bf \ast \boldsymbol{\psi}_j = & \sum_{l=0}^{N-1} \Bu_l \Bu_l^* \Bf \hat{\psi}(2^{-j} \l_l), \text{ for } j>-J ~; \\
\Bf \ast \boldsymbol{\phi}_{-J} = & \sum_{l=0}^{N-1} \Bu_l \Bu_l^* \Bf \hat{\phi}(2^{J} \l_l)~. 
\end{aligned}
\end{equation*}

For any $m$ no larger than the number of layers, a path $p = (j_1, \cdots, j_m)$ is a vector of $m$ scales of the graph wavelets, which satisfy $0 \leq j_1, \cdots, j_m \leq J-1$. The scattering propagator with respect to a path $p$ is obtained by consecutive application of convolutions with wavelets of these scales and absolute values, which serve as nonlinearities, in the following way 
\begin{equation*}
    \BU[p] \Bf = \abs{ \abs{ \abs{\Bf \ast \Bpsi_{j_1}} \ast \Bpsi_{j_2} } \ast \cdots \ast \Bpsi_{j_m} } ~.
\end{equation*}
The scattering transform with respect to the path $p$ is obtained by 
$\BS[p] \Bf  = \BU[p] \Bf \ast \Bphi_{-J}$.

Let $\cP$ denote the collection of all paths of length no larger than the number of layers. The scattering transform of $\Bf$ with respect to $\cP$ is
\begin{equation*}
    \BS[\cP] \Bf = ( \BS[p] \Bf )_{p \in \cP}.
\end{equation*}
A simple illustration of the scattering transform is provided in Fig. 1 of \cite{zou2018graph}. Note that the scattering transform depends on the underlying graph. For simplicity, we exclude this dependence from our notation.

For the $K$-dimensional signal $\BX = [\BX_1 | \cdots | \BX_K] \in \RR^{N \times K}$, the scattering transform is
\begin{equation*}
    \BS[\cP] \BX = (\BS[\cP] \BX_k)_{k=1}^K ~.
\end{equation*}
We remark that if the set ${\cP}$ has $L$ elements, then defining $M = L K$, $\BS[\cP](\BX)$ can be represented as a matrix in $\RR^{N \times M}$ or a vector in  $\RR^{N \cdot M}$. We denote the latter vector by $\bar{\BX}$.

Zou and Lerman \cite{zou2018graph} establish various theoretical properties of the scattering transform $\BS[\cP]$. In particular, they show that it preserves the energy---that is, the $l_2$ norm---of each signal. The usefulness of this property for graph generation will be discussed later. Another important property they establish is the robustness of the scattering transform to small perturbations of the signal and the graph \cite[\S 5]{zou2018graph}. This robustness implies that similar signals and graphs are encoded as similar latent codes. 

The last step of the encoder maps the transformed data points so that one may possibly generate them with a Gaussian distribution, as required by the generator. We earlier referred to this mapping as Gaussianization. We describe two possible mappings that were previously suggested for related tasks. Following \cite{angles2018generative}, one such mapping is whitening. Specifically, let
$\{\bar{\BX}^{(t)} \}_{t=1}^T$  be the scattering transform vectors of the input samples and let $\bar{\boldsymbol{\mathcal{X}}}$ be the representing matrix of $\{\bar{\BX}^{(t)} \}_{t=1}^T$. That is, $\bar{\boldsymbol{\mathcal{X}}} = (\bar{\BX}^{(t)})_{t=1}^{T} \in \RR^{T \times NM}$. As advocated in  \cite{angles2018generative}, a dimension reduction by PCA can be further applied to $\{\bar{\BX}^{(t)} \}_{t=1}^T$. Next, using the following mean and sample covariance of the scattering transform vectors
\begin{equation*}
    \Bmu = \frac{1}{T} \sum_{t=1}^T \bar{\BX}^{(t)}  \ \text{ and } \
    \BSigma = \frac{1}{T} \sum_{t=1}^T (\bar{\BX}^{(t)} - \Bmu) (\bar{\BX}^{(t)} - \Bmu)^* ~,
\end{equation*}
the whitening map $\BA$ is 
\begin{equation}\label{eq:A}
    \BA \bar{\boldsymbol{\mathcal{X}}} = \BSigma^{-1/2} (\bar{\boldsymbol{\mathcal{X}}} - \Bmu) ~.  
\end{equation}
The whitened samples are uncorrelated and the hope in \cite{angles2018generative} is that their distribution will be close to that of a normal distribution with identity covariance. 

The output of the encoder with a whitening transformation is guaranteed to have similar robustness results to signal and graph manipulations as the ones established in \cite{zou2018graph} for the graph scattering transform with additional factor of $\|\BSigma^{-1/2}\|$, where $\|\cdot\|$ denotes the spectral norm of the corresponding matrix. We remark that due to the initial dimension reduction that removed small eigenvalues of the sample covariance of the transformed data points, $\|\BSigma^{-1/2}\|$, is not expected to be large.

One problem with the whitening process is that the distribution of its output may not be close to Gaussian. While traditional Gaussianization methods \cite{chen2001gaussianization, laparra2011iterative} can improve the distribution of the output, their encoding may not be robust to signal and graph manipulation.

An alternative to whitening is a spherization procedure inspired by \cite{bojanowski2018optimizing}.
That is, the data points $\{\bar{\BX}^{(t)} \}_{t=1}^T$ are normalized to lie on the unit sphere in the Euclidean norm. The hope is that these mapped data points can be generated from a Gaussian with their sample mean and covariance. In general, this may not be the case, though \cite{bojanowski2018optimizing} have a heuristic and incomplete argument for this property in a different setting. This argument {can be made precise by using} the Gaussian Annulus Theorem (see e.g., \cite[Theorem 2.9]{blum2017foundation}). 

Due to the energy preservation property of the full graph scattering transform of \cite{zou2018graph}, one can instead spherize the input data points and the result of the encoder will be the same. Therefore, in theory, the encoder is robust to signal and graph manipulation with respect to the original data after spherization. Due to this property and similarly to \cite{bojanowski2018optimizing}, we do not initialize the spherization by centering with the sample mean. In practice, since scattering is only applied to a finite number of layers, the energy is contracted. Therefore, it is better {to} normalize after the scattering and this is what we do in our experiments. The final stage of the encoder with spherization calculates the sample mean $\Bmu_s$ and covariance $\Bsigma_s$ for the spherized scattering output and fits a Gaussian $\cN(\Bmu_s, \Bsigma_s)$, which is used as the latent distribution for sampling.

We denote the original samples by $\{{\BX}^{(t)} \}_{t=1}^T$  and the corresponding data matrix in $\RR^{T \times NM}$  by ${\boldsymbol{\mathcal{X}}}$.
We further denote the output of the encoder (with either whitening or spherization) by $\BPhi [\cP] (\boldsymbol{\mathcal{X}})$, or in short $\BPhi(\boldsymbol{\mathcal{X}})$. We note that the mapping $\BPhi$ can also be applied to any signal $\BX \in \RR^{N \times K}$. We denote the feature vector corresponding to the $K$-dimensional signal $\BX$ by $\Bz  = \BPhi[\cP](\BX) \in \RR^{N \times M}$. We will refer to $\Bz$ as a latent code.

\subsection{Details of the decoder}
\label{sec:graph_scat_decode}
Recall that the decoder is a network with fully-connected layers. We describe its architecture according to the following three different tasks.  

\subsubsection{Link prediction}\label{subsec:theolinkpred}
\begin{figure}[t]
    \centering
    \includegraphics[width=.9\linewidth]{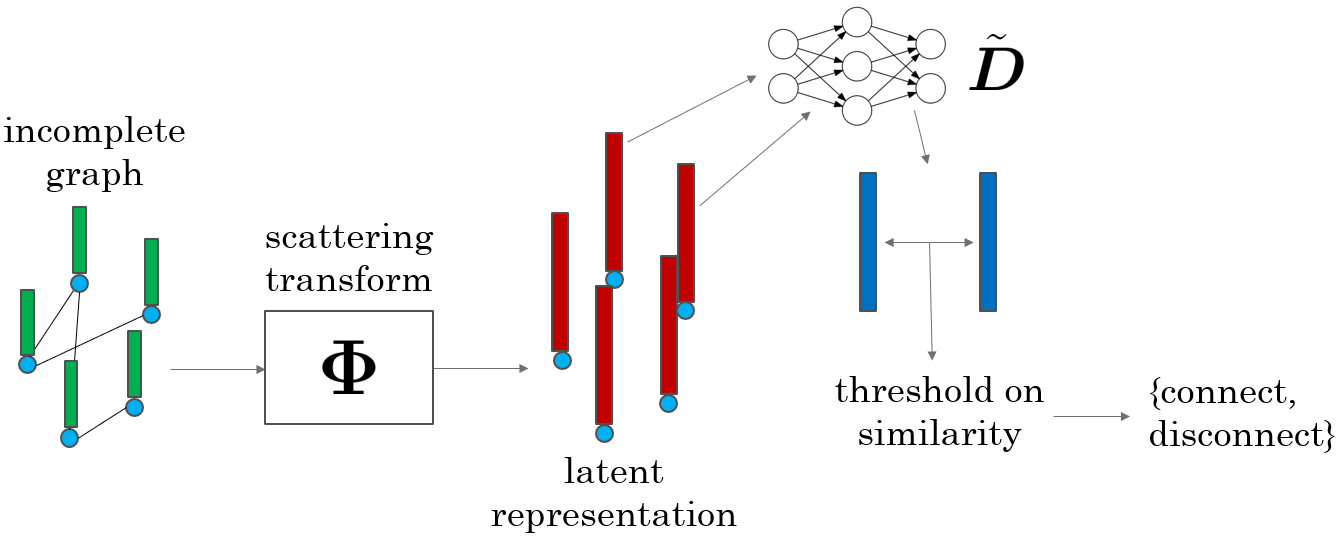}
    \caption{Sketch of a graph scattering network for link prediction.} 
    \label{fig:linkpred}
\end{figure}

For link prediction, we encode the features of the partially available graph into a latent vector, and use the same vector to generate the full graph via the learned decoder. Note that in this task only one fixed graph is given, and thus no Gaussianization procedure is applied in the encoder. That is, the linear transformation $\BA$ in \eqref{eq:A} is the identity. The input includes a weight matrix $\BW_{\textup{train}}$, which contains weights for the partially available edges, and a feature matrix $\BX \in \RR^{N \times K}$ of $K$-dimensional signals on the $N$ nodes. The encoder is a scattering network $\BPhi$ that maps $\BX$ and $\BW_{\textup{train}} \in \RR^{N \times N}$ into a latent representation $\Bz \in \RR^{N \times M}$. As in \cite{kipf2016variational}, the decoder is a simple network $\BD$ such that $\BD(\Bz) = \sigma(\tilde{\BD}(\Bz) \tilde{\BD}(\Bz)^T)$, where $\sigma$ is the softmax function 
and $\tilde{\BD}$ is an MLP. The network $\BD$, whose parameters are those of $\tilde{\BD}$, is trained to minimize the 
cross-entropy loss function 
\begin{equation}\label{eq:crossentropylink}
    L(\BD) = \sum_{i, j: \BW(i,j) \neq 0} [- \log \BD(\BPhi(\BX,\BW)) (i,j)] ~.
\end{equation}
The structure of the entire network is illustrated in Fig. \ref{fig:linkpred}.

\subsubsection{Signal generation on graphs}\label{subsec:theographsignalgen}
\begin{figure}[t]
    \centering
    \includegraphics[width=.9\linewidth]{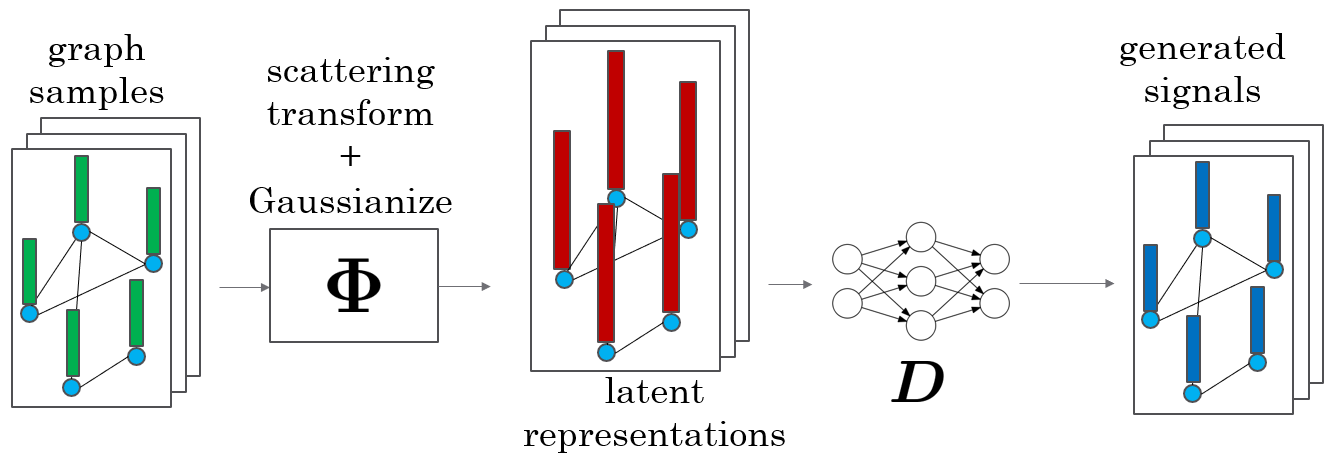}
    \caption{Sketch of a graph scattering network for signal generation on graphs.} 
    \label{fig:signal}
\end{figure}

For signal generation on graphs, which we also refer to as graph signal generation, one is given a fixed graph domain and different signals on the nodes of this graph and the goal is to generate similar signals. An input random variable $\BX \in \RR^{N \times K}$ is first mapped by the scattering transform to $\BS[\cP](\BX)$ and then to a Gaussian $\Bz = \BPhi[\cP](\BX)\in \RR^{N \times M}$. The decoder $\BD$ is taken to be an MLP that maps $\Bz$ to a matrix $\BD(\Bz)$ in $\RR^{N \times K}$. The parameters of $\BD$ are obtained by minimizing the reconstruction loss function 
\begin{equation}\label{eq:lossgraphsignalgen}
    L(\BD) = T^{-1} \sum_{t=1}^T \norm{\BX^{(t)} - \BD(\BPhi(\BX^{(t)}))} ~, 
\end{equation}
where $\{ \BX^{(t)} \}_{t=1}^T$ are the training data points.
The structure of the generative network is the same as in Fig. \ref{fig:mallat}, where in the current case $S$ is the graph scattering transform.
Fig. \ref{fig:signal} illustrates the generation procedure.

\subsubsection{Graph and signal generation}\label{subsec:theographandsignalgen}
\begin{figure}[t]
    \centering
    \includegraphics[width=.9\linewidth]{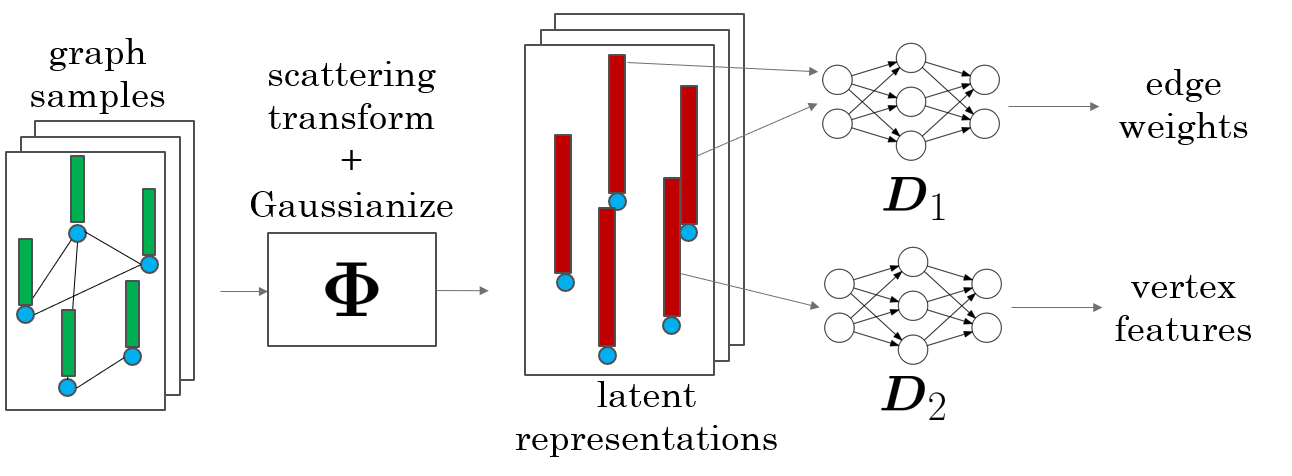}
    \caption{Sketch of the graph scattering network for graph and signal generation.} 
    \label{fig:chem}
\end{figure}

The scattering transform can be used as an encoder for generating both the graph and the signal on it. We train two MLP's $\BD_1$ and $\BD_2$, where both take the Gaussian random variable $\Bz = \BPhi(\BX)$ as input. The network $\BD_1$ is used to generate the graph signal $\BX$ and the network $\BD_2(\Bz) = \sigma(\tilde{\BD}_2(\Bz) \tilde{\BD}_2(\Bz)^T)$ is used to generate the weight matrix $\BW$. They are trained at the same time, with the reconstruction loss function 
\begin{equation}\label{eq:lossgraphandsignal}
\begin{aligned}
 L(\BD_1, \BD_2) = & T^{-1} \sum_{t=1}^T \Big[ \norm{\BW^{(t)} - \BD_1(\BPhi(\BX^{(t)}))} + \\
  & \norm{\BX^{(t)} - \BD_2(\BPhi(\BX^{(t)}))} \Big],
\end{aligned}
\end{equation}
where $\{ ( \BX^{(t)}, \BW^{(t)} ) \}_{t=1}^T$ are the training data points. The norms can be replaced with cross-entropy losses if one wants the outputs to be categorical, {as in the molecule generation task}.
Fig. \ref{fig:chem} illustrates this generation procedure. 

\section{Experiments}\label{sec:experiments} 
We test our proposed method, whose code is available at https://github.com/dmzou/SCAT, and compare it with other available algorithms using datasets addressing the three different tasks reviewed in \S\ref{sec:intro}.

Even though we advocate using the graph scattering network of \cite{zou2018graph} (due to its robustness to signal manipulation), we also tested our generation algorithm with the graph scattering network of \cite{gama2018diffusion}. When addressing link prediction (so a Gaussianization procedure is not applied), we denote by SCAT-S and SCAT-D our proposed generative procedure with the spectral and diffusion networks of \cite{zou2018graph} and \cite{gama2018diffusion} respectively. For the other two applications, we distinguish between the two Gaussianization procedures: whitening and spherization. 
We use ``W'' to denote whitening and ``N'' to denote normalization, i.e., spherization. We use, as above, ``S'' to denote the spectral graph scattering transform in \cite{zou2018graph} and ``D'' to denote the diffusion graph scattering transform in \cite{gama2018diffusion}. We denote the four resulting networks by SCAT-SW, SCAT-DW, SCAT-SN and SCAT-DN. 

Following \cite{zou2018graph} we use in all experiments the simple Shannon wavelet and {the limiting scale} $J = 3$ for SCAT-S, SCAT-SW and SCAT-SN. We similarly use $J = 3$ and choose $t=3$ ($t$ is the power of multiplying diffusion operator) for SCAT-D, SCAT-DW and SCAT-DN. For link prediction, we take 2-layer scattering as it performs better on the validation set; for the other two tasks, we take 3-layer scattering.

Our method requires moderate memory and GPU and we easily tested it on a PC with 8GB RAM and GTX1060 GPU. Comparison with VGAE on the Pubmed dataset demanded more advanced GPU. We thus report results of all experiments on a Linux machine with 64GB RAM and GTX1080Ti GPU.

\subsection{Link prediction for citation data}\label{subsec:linkpred}

We predict links for the three citation datasets of \cite{sen2008collective}: Cora, Citeseer and Pubmed. Each dataset contains information about publications in  certain fields and the corresponding citation linkage between these publications. This information can be embedded in a graph in which the publications and citations are represented by vertices and edges, respectively. Although the citation link is directed, we follow the common convention of assuming an undirected and unweighted graph. That is, if any of two papers cite the other, an edge is drawn between them. 
Table \ref{tab:dataset} lists some characteristics of the three datasets. 

\begin{table}[t]
\caption{Characteristics of three citation datasets}\label{tab:dataset}
\centering
\begin{tabular}{lrrrr}
Dataset    &    Vertices    &    Edges    &    Classes    &    Features \\
\hline
Cora     &    2708    &    5429    &    7        &    1433    \\
Citeseer    &    3327    &    4732    &    6        &    3703    \\
Pubmed    &    19717    &    44338    &    3        &    500    \\
\end{tabular}
\end{table}

We use the preprocessing step suggested in \cite{kipf2016variational} and documented in  {https://github.com/tkipf/gae} for all tests. It divides the original edges into 85\%  training,  5\% validating and 10\% testing sets. 

For SCAT-S and SCAT-D, the dimension of the output of the scattering transform is reduced to 128. 
The decoder $\tilde{\BD}$ in both models is taken to be a single dense layer of size 512, activated by the ReLU function.   In order to minimize \eqref{eq:crossentropylink}, we use the Adam optimizer \cite{kingma2014adam} with a learning rate of 0.001, where we train 1,000 epochs for each run. We record the following two common scores: area under curve (AUC) and average precision (AP).  Similarly to \cite{kipf2016variational}, we take 10 runs for each setting and record the average and standard deviation. 
Table \ref{tab:linkprediction} reports our results for SCAT-S and SCAT-D, together with results for GAE and VGAE \cite{kipf2016variational}.
We note that SCAT-S improves over the previous results for all the three datasets, while SCAT-D achieves comparable results with GAE and VGAE. The averaged time for the scattering transform of SCAT-S and SCAT-D, applied to Cora/Citeseer/Pubmed, is {2.82s/6.43s/492.90s} and {1.77s/4.31s/346.43s} respectively. SCAT-S requires more time due to the spectral decomposition, whose average time for the 3 datasets is {1.24s/2.16s/317.43s}. Table \ref{tab:epochlink} reports the training time for each epoch. This table implies that SCAT-S and SCAT-D are much more efficient. This is because only parameters in the decoder need to be updated. 
Combining the times of the scattering transforms reported 
above, which are executed once, and the training times for each epoch in Table \ref{tab:epochlink}, multiplied by the number of epochs, we conclude that the total times of SCAT-S and SCAT-D are more efficient.

\begin{table*}[ht]
\centering
\caption{Results for link prediction using the citation datasets. We report the mean and standard deviation over 10 runs for each setting. All models for the same dataset are trained based on the same training and validating links.} 
\small
 \begin{tabular}{| l | c c | c c | c c |} 
 \hline
 Dataset & \multicolumn{2}{|c|}{Cora} & \multicolumn{2}{|c|}{Citeseer} & \multicolumn{2}{|c|}{Pubmed} \\
 \hline
  & AUC (\%) & AP (\%) & AUC (\%)  & AP (\%) & AUC (\%)  & AP (\%)  \\ [0.5ex] 
 \hline
SCAT-S & 94.48 $\pm$ 0.15 & 94.63 $\pm$ 0.17 &  97.27 $\pm$ 0.12 & 97.57 $\pm$ 0.12 &  97.52 $\pm$ 0.03 & 97.19 $\pm$ 0.04  \\
 \hline
SCAT-D & 92.08 $\pm$ 0.09 & 93.05 $\pm$ 0.11 &  92.54 $\pm$ 0.14 & 94.16 $\pm$ 0.12 &  92.73 $\pm$ 0.17 & 93.56 $\pm$ 0.09  \\
 \hline
 GAE  &  91.34 $\pm$ 0.52 & 92.62 $\pm$ 0.38 &  92.37 $\pm$ 0.67 & 93.72 $\pm$ 0.58 & 96.35 $\pm$ 0.18 & 96.53 $\pm$ 0.16 \\
 \hline
VGAE & 91.14 $\pm$ 0.40 & 92.16 $\pm$ 0.29 &  92.70 $\pm$ 0.76 & 93.93 $\pm$ 0.57 &  95.68 $\pm$ 0.35 & 95.92 $\pm$ 0.32 \\
 \hline
\end{tabular}
\label{tab:linkprediction}
\end{table*}

\begin{table}[ht]
\caption{Time for training an epoch for the citation data}\label{tab:epochlink}
\centering
\begin{tabular}{lrrrr}
Dataset    &   SCAT-S    &    SCAT-D    &    GAE & VGAE    \\
\hline
Cora    &   8.1ms    &  8.1ms     &      209.1ms & 206.4ms      \\
Citeseer     &  8.1ms     &  8.1ms    &   298.6ms   & 302.3ms  \\
Pubmed    &  64.9ms     &  64.9ms     &  7832.6ms    &  7889.2ms   \\
\end{tabular}
\end{table}

\subsection{Signal generation on graphs}\label{subsec:graphsignalgen}

We use the Fashion-MNIST dataset \cite{xiao2017fashion} for a sanity check of SCAT for the problem of graph signal generation. 
Any element of this dataset is a $28 \times 28$ grayscale pixel image and can be considered as a graph in the following way: the pixels are the graph vertices, and nearby pixels are connected by graph edges. 
The edges and weights are formed as in \cite{zou2018graph}. That is, each pixel represents a vertex and it is connected with its four nearest neighbors with weight $e^{-1}$ and its four nearest diagonal neighbors with weight $e^{-2}$.

The encoder of this network is a graph scattering transform. Its output dimension is reduced to 256 from 28 $\times$ 28 $\times$ 13 = 10,192. The decoder is an MLP of two hidden layers of size 512. In order to minimize \eqref{eq:lossgraphsignalgen}, we use the Adam optimizer with a learning rate of 0.001, where we train 2,000 epochs for each run. {To avoid mode collapse,} we have restricted the dataset to the ``boots'' category, which contains 5,454 training examples. Sample images from this category are demonstrated in Fig. \ref{fig:fashionorigin}.

To generate an ``image'' (or graph signal), we take a sample $\Bz \in \RR^N$ from $\cN(0,I)$ for SCAT-SW and SCAT-DW, and from $\cN(\Bmu_s, \BSigma_s)$ for SCAT-SN and SCAT-DN, and report the output of the decoder. 
Figs. \ref{fig:fashion-scat-sw}-\ref{fig:fashion-scat-dg} illustrate samples generated by these networks, while using the graph associated with a $28 \times 28$ pixel image. We report the $\ell_1$ reconstruction loss defined in \eqref{eq:lossgraphsignalgen} in Table \ref{tab:epochsignal}. Note that the averaged $\ell_1$-norm of the training images is $174.56$. We also report the times for scattering and training a single epoch. 

\begin{table}[b]
\caption{Reconstruction loss and time for scattering transform (for the complete training data) and training an epoch for the Fashion-MNIST dataset}\label{tab:epochsignal}
\centering
\begin{tabular}{lrrrr}
~    &   SCAT-SW    &    SCAT-DW    &    SCAT-SN & SCAT-DN    \\
\hline
Reconstr. loss   &   0.0482    &  0.0532     &      0.0513 & 0.0548     \\
Time (scattering)     &  1,128.6ms     & 1,484.3ms     &   1,148.3ms & 1,566.0ms      \\
Time (epoch)      &  39.4ms     &  39.1ms     &    30.7ms & 30.6ms       \\
\end{tabular}
\end{table}

\begin{figure}[ht]
    \centering
    \includegraphics[width=.67\linewidth]{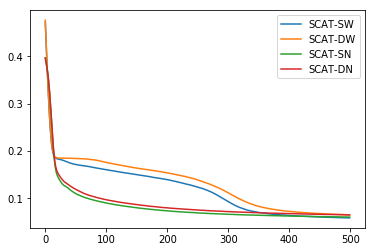}
    \caption{Reconstruction loss with respect to epoch for training SCAT models for the Fashion-MNIST data. No difference is noted after 500 epochs.} 
    \label{fig:loss}
\end{figure}

\begin{figure*}[t]
  \centering
  \begin{subfigure}[b]{.24\linewidth}
    \centering\includegraphics[width=\linewidth]{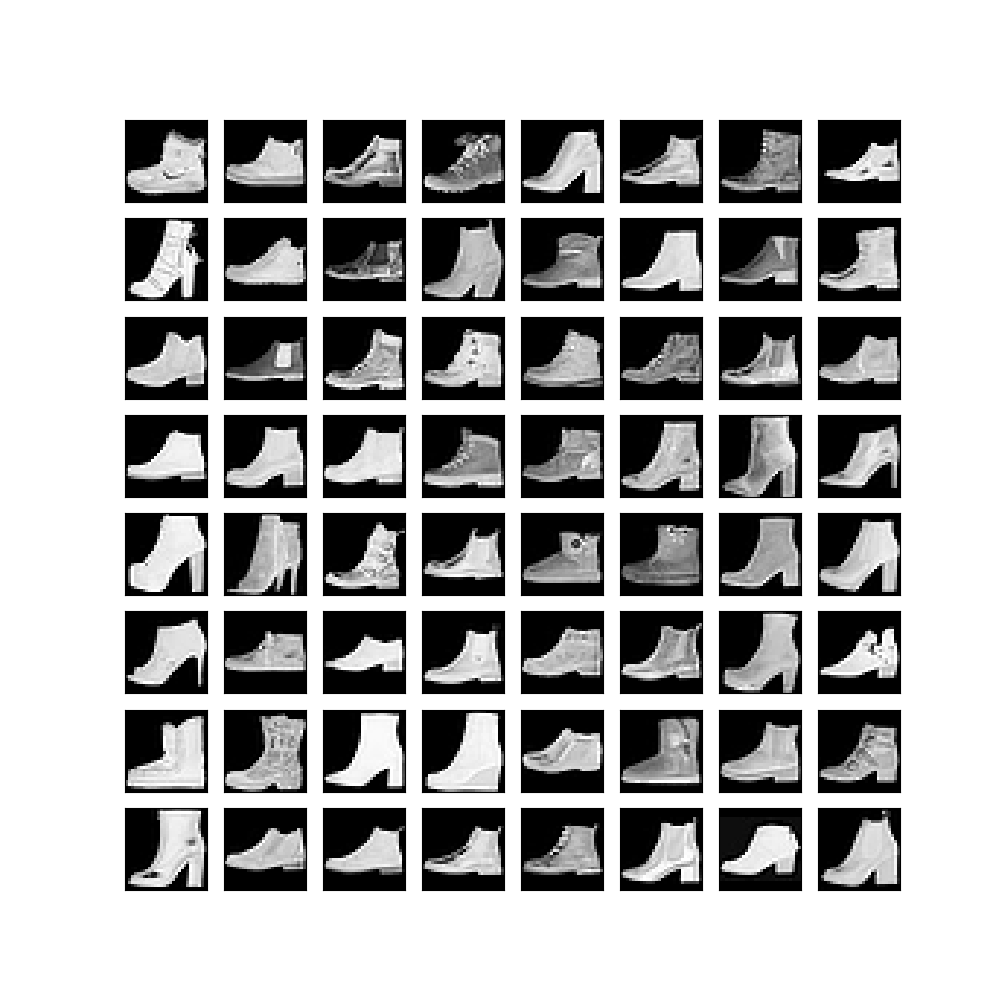}
    \caption{Original data. 
    \label{fig:fashionorigin}}
  \end{subfigure}%
  \begin{subfigure}[b]{.24\linewidth}
    \centering\includegraphics[width=\linewidth]{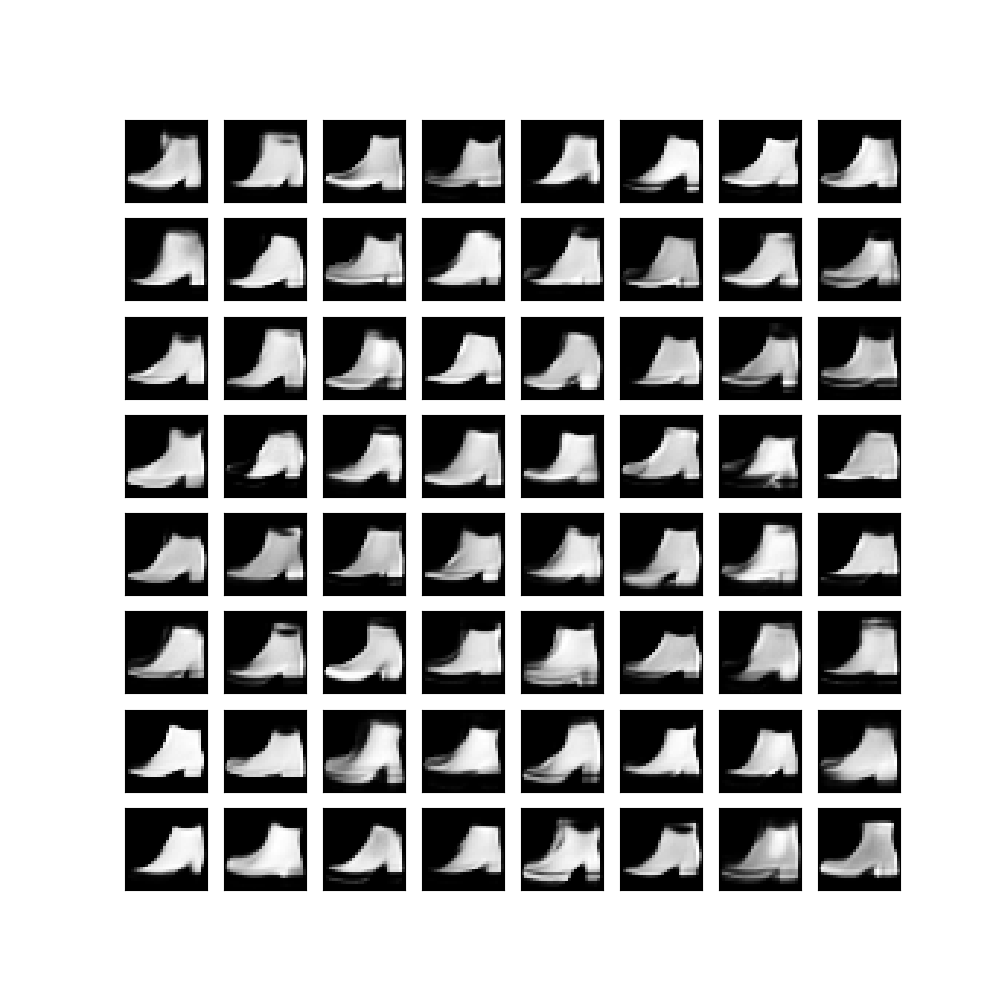}
    \caption{SCAT-SW.  
  \label{fig:fashion-scat-sw}}
  \end{subfigure}
  \begin{subfigure}[b]{.24\linewidth}
    \centering\includegraphics[width=\linewidth]{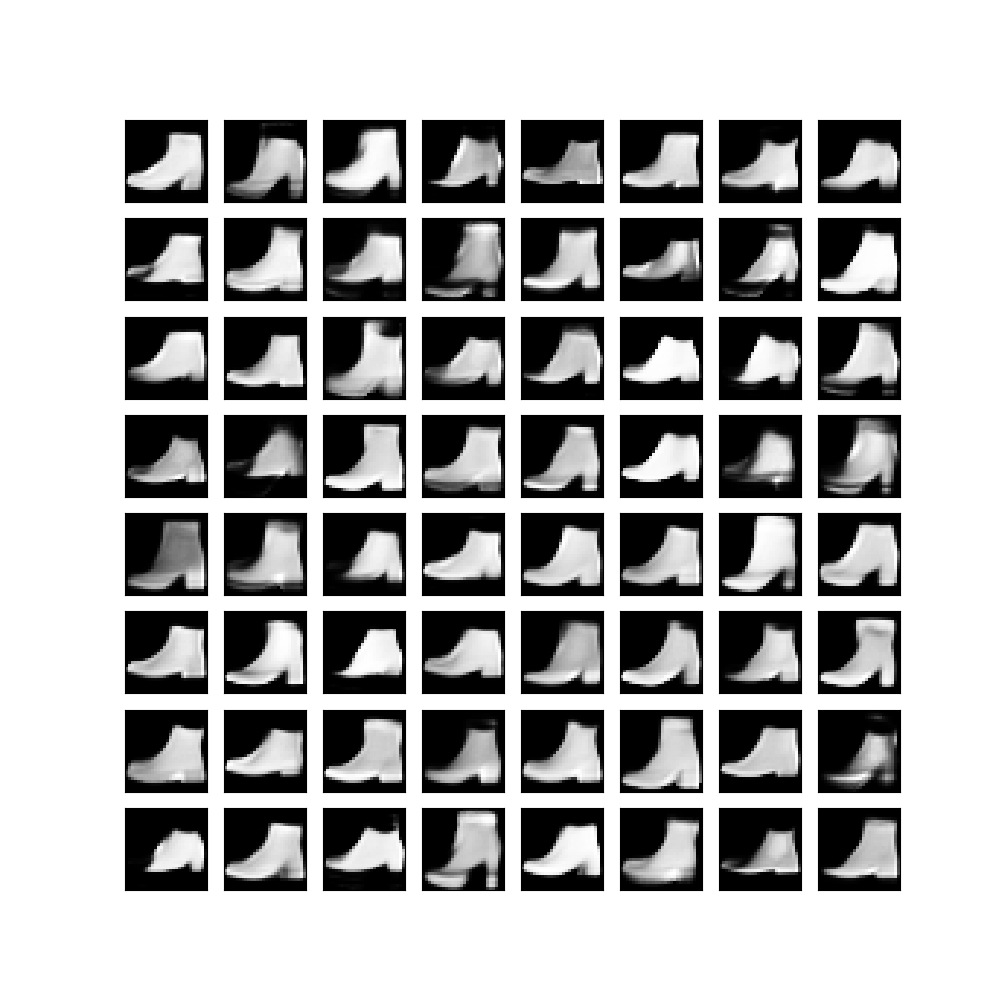}
    \caption{SCAT-SN. 
    \label{fig:fashion-scat-sg}}
  \end{subfigure}
  \begin{subfigure}[b]{.24\linewidth}
    \centering\includegraphics[width=\linewidth]{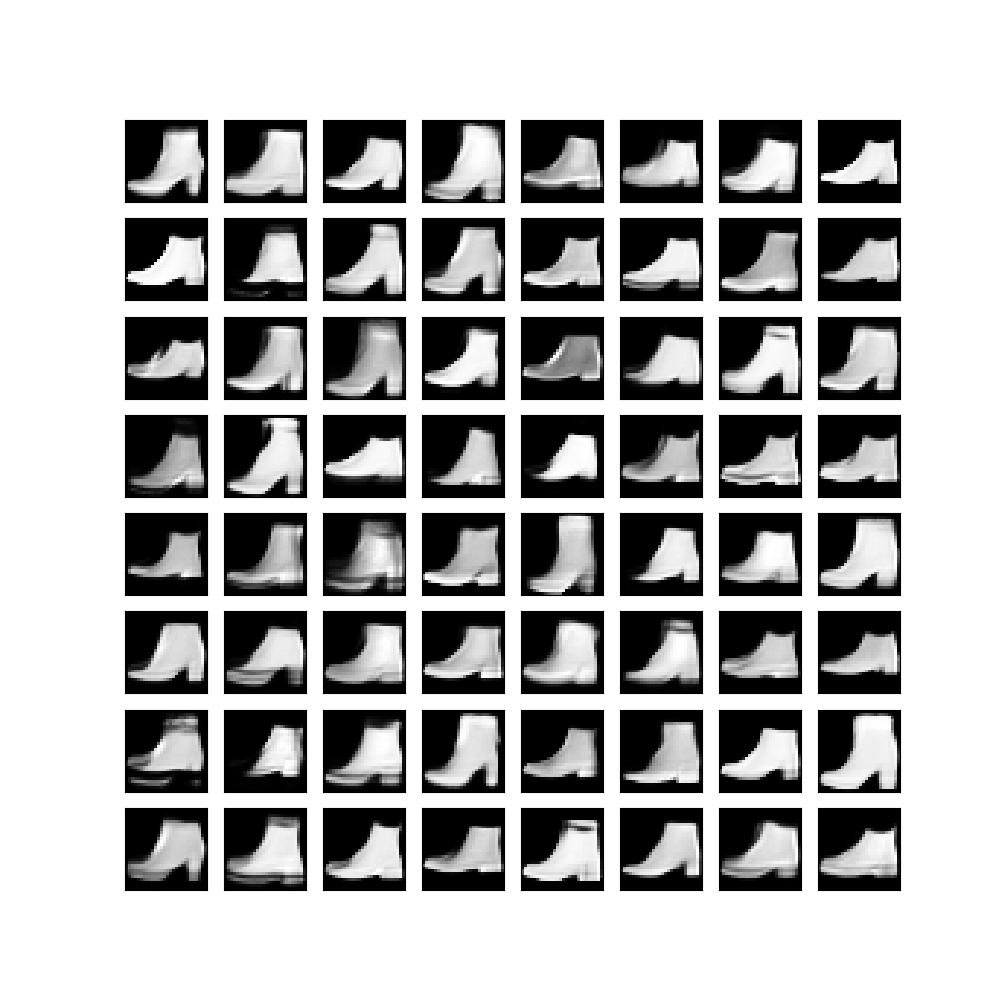}
    \caption{SCAT-DW. 
    \label{fig:fashion-scat-dw}}
  \end{subfigure}
  \hfill
  \begin{subfigure}[b]{.24\linewidth}
    \centering\includegraphics[width=\linewidth]{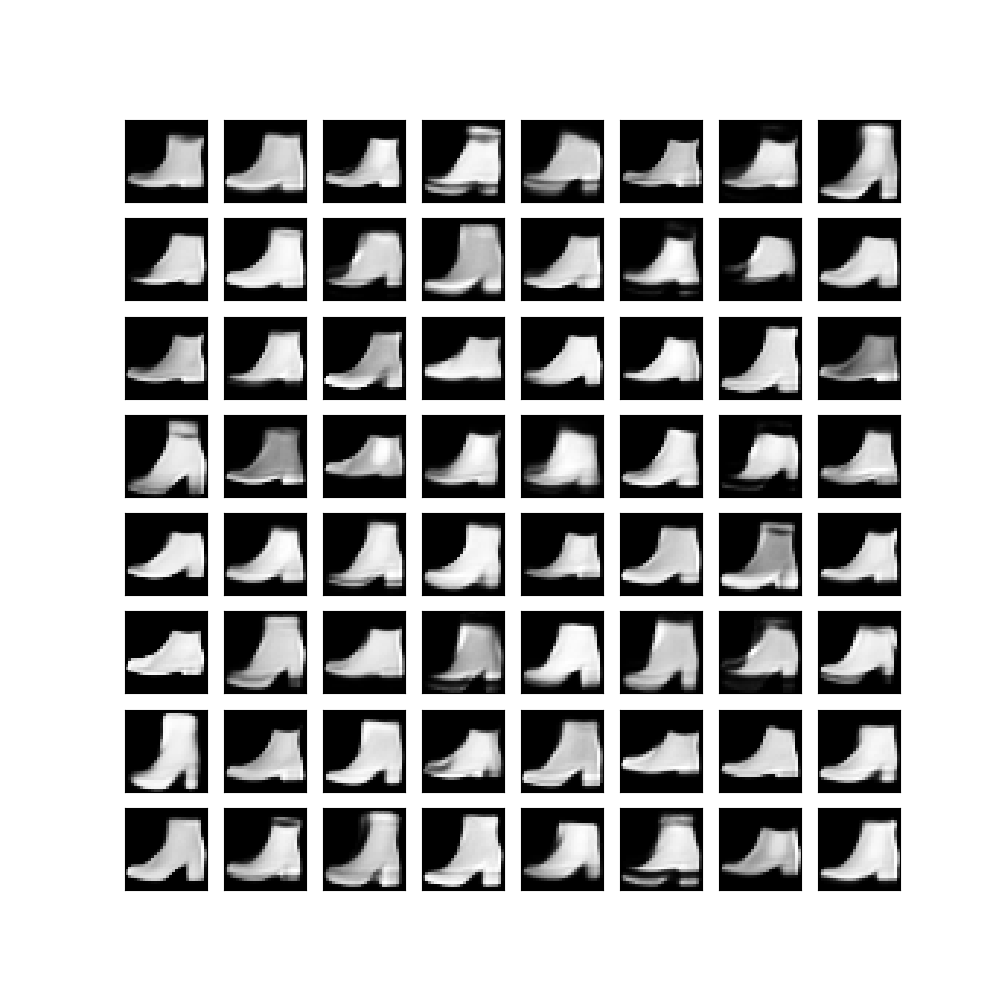}
    \caption{SCAT-DN. 
    \label{fig:fashion-scat-dg}}
  \end{subfigure}%
  \begin{subfigure}[b]{.24\linewidth}
    \centering\includegraphics[width=\linewidth]{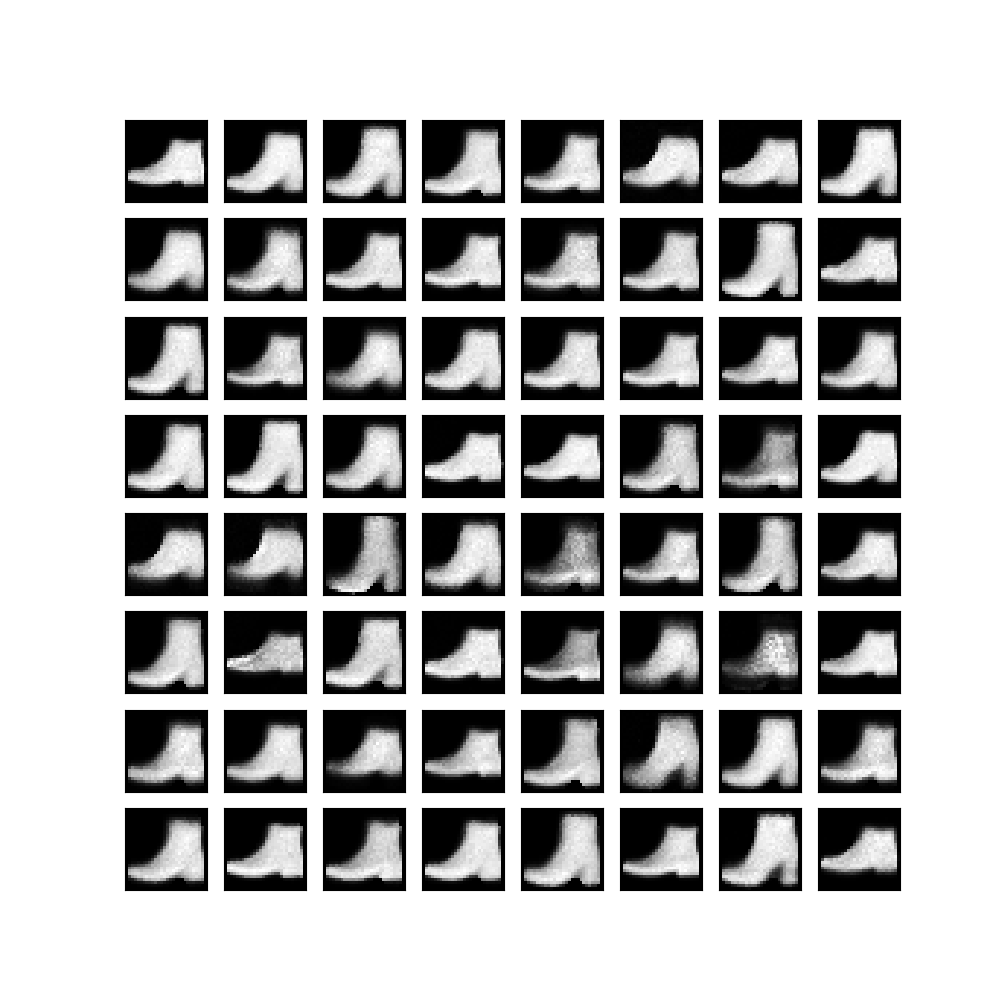}
    \caption{VAE-GCN.  
  \label{fig:fashion-vae}}
  \end{subfigure}
  \begin{subfigure}[b]{.24\linewidth}
    \centering\includegraphics[width=\linewidth]{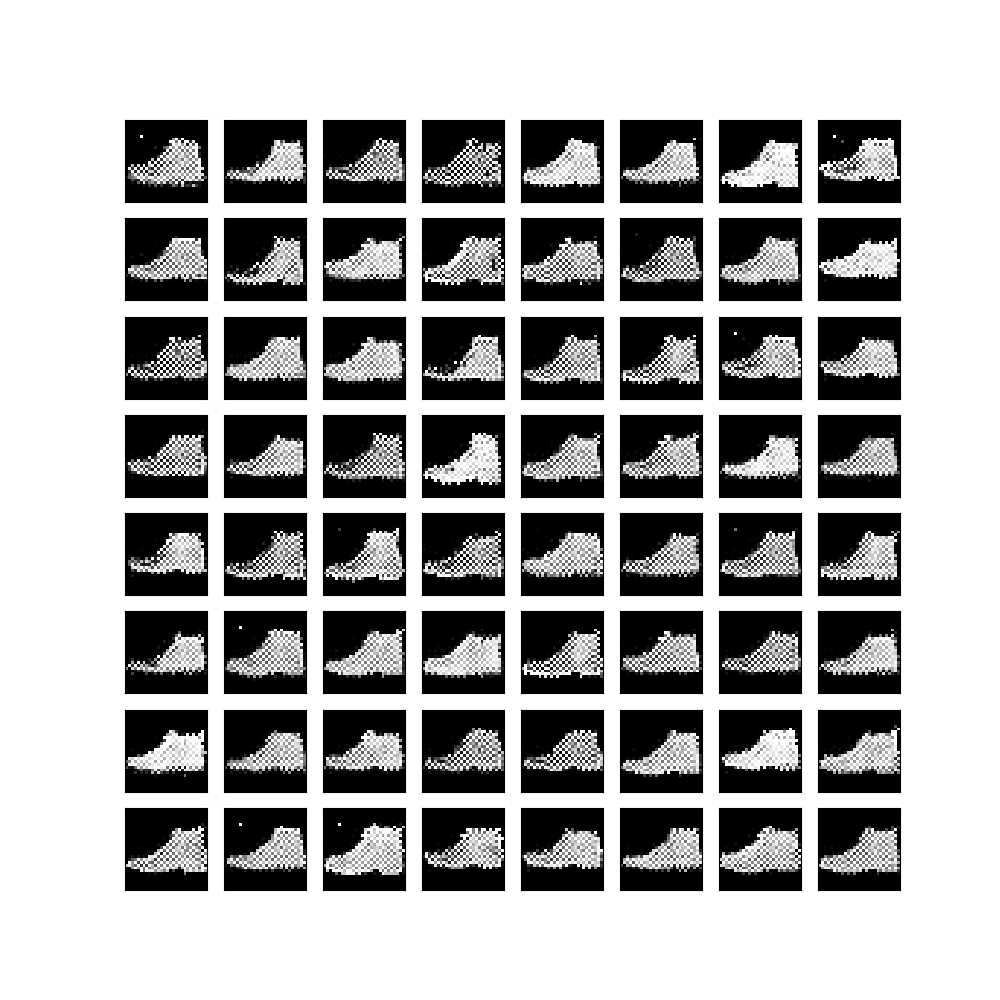}
    \caption{GAN-GCN. 
    \label{fig:fashion-gan-gnn}}
  \end{subfigure}  
  \caption{Original and generated images for boots of Fashion-MNIST data.}
  \label{fig:mnist}
\end{figure*}

Our experiments indicate that training models with spherization converges faster than with whitening. To see this, we plot the reconstruction loss as a function of epoch for the first 500 epochs for the four methods we tested in Fig. \ref{fig:loss}. It is clear that SCAT-SN and SCAT-DN converge much faster in the first 300 epochs.

The images generated in Figs. \ref{fig:fashion-scat-sw}-\ref{fig:fashion-scat-dg} are of similar quality for all the scattering methods. Note that some high frequency components, that is, details, of these images are missing. This phenomena is is common with the Euclidean generative scattering network of \cite{angles2018generative}. Nevertheless, common graph-type data often do not have high-frequency {components}. For instance, in the molecular data reviewed in \S\ref{subsec:graphgen}, the signals on each vertex just take five different values. 

For comparison with other methods, Figs. \ref{fig:fashion-vae} and \ref{fig:fashion-gan-gnn} illustrate samples generated from graph generative models that combine VAE/GAN with the graph convolutional layers proposed in \cite{kipf2016semi}. Specifically, for what we call VAE-GCN, we construct a graph-based VAE in which the encoder consists of two graph convolutional layers while the decoder is an MLP of two hidden layers. The latent mean and variance both have dimension 256 and the hidden layers of the MLP have dimension 512. For what we call GAN-GCN, we replace the discriminator of a vanilla GAN with two graph convolutional layers and use an MLP with two hidden layers of dimension 512 for the generator. 
  
All three methods are trained using Adam with a learning rate of 0.001. The latent noise for both GAN-GCN and GAN-FCN is of dimension 256. The training time for each epoch for VAE-GCN and GAN-GCN is 3.947s and 1.761s respectively, which is much slower than training SCAT models since parameters from both the generator and the discriminator need to be updated for GAN-GCN, and parameters from both the encoder and the decoder need to be updated for VAE-GCN. Observing Figs. \ref{fig:fashion-vae} and \ref{fig:fashion-gan-gnn}, we see that samples from VAE-GCN look blurrier than those of SCAT models and also miss high-frequency information. Samples from GAN-GCN are very noisy and still miss high-frequency information and suggest that the model suffers from a severe mode collapse.

\subsection{Graph generation for molecular data}\label{subsec:graphgen}
We test graph and signal generation using the molecular dataset QM9 \cite{ramakrishnan2014quantum}. This dataset contains 134k molecules made of the following atoms: C, H, O, N, and F. There are two common ways of embedding these kinds of datasets into an interpretable feature space. Kusner et al.~\cite{kusner2017grammar} treat molecules as ``words'' by looking at their simplified molecular-input line-entry (SMILE) strings.
Graphs are also commonly used to represent molecules, where the graph vertices represent the atoms composing the molecule and the graph edges are the bonds. The vertex signals assign the four different atom types to the vertices. While there are five type of atoms, H is automatically determined by the other atoms and the given chemical bonds. Therefore, only labels of the four heavy atoms (C, O, N or F) need to be assigned. Since the dataset has molecules with at most 9 heavy atoms, we assume graphs with 9 vertices and assign a dummy value for vertices without assignment of a heavy atom.
Each heavy atom and the dummy one are assigned a one-hot vector, i.e., a unit coordinate vector in $\RR^5$, and each atom is represented by one of these five one-hot vectors.

For graph generation, there is no unique benchmark for checking the quality of the generated graphs. Bojchevski et al.~\cite{bojchevski2018netgan} proposed to use graph properties such as the max degree and the number of triangles for graph generation. However, it is often hard to compare these graph properties and there is no motivation for using them for molecule generation.
Samanta et al.~\cite{samanta2018designing} proposed to check validity (whether a sample is a valid chemical molecule), uniqueness (whether a sample is unique among all generated samples) and novelty (whether a sample is different from any sample in the training data). We use these measures since they are more quantitative and experiments on QM9 by \cite{de2018molgan} and \cite{simonovsky2018graphvae} also report them. They are checked after converting the graphs into SMILE strings using the RDKit package (https://www.rdkit.org/).

The full QM9 dataset is used for training. This choice is the same as that in \cite[\S 5.3]{de2018molgan}. It is different than  \cite{simonovsky2018graphvae}, in which a small set of molecules is used for training and only molecules with 9 heavy atoms are considered. 

As explained in \S\ref{subsec:theographandsignalgen}, for this application of graph generation, the SCAT decoders for both vertices and edges are MLP's. In our experiments, both of them have three hidden layers of dimension 128, 256, 512, respectively. We take the encoder to be a graph scattering transform, with output dimension reduced to 135 (15 for each vertex).
In order to minimize \eqref{eq:lossgraphandsignal}, we use the Adam optimizer with a learning rate of 0.001, where we train 300 epochs.
Table \ref{tab:epochmol} reports the computational times of scattering and training on this dataset.

\begin{table}[t]
\caption{Time for the scattering transform (for the complete training data) and training an epoch for the QM9 dataset}\label{tab:epochmol}
\centering
\begin{tabular}{lrrrr}
~    &   SCAT-SW    &    SCAT-DW    &    SCAT-SN & SCAT-DN    \\
\hline
Time (scattering)     & 137.12s     &  94.75s    &   132.01s & 95.95s      \\
Time (epoch)      &  3.68s     &  3.66s     &    3.71s & 3.69s       \\
\end{tabular}
\end{table}

Using SCAT, we generate 10k molecules and record the validity, uniqueness, and novelty in Table \ref{tab:molcpr}. For comparison, we also record results reported for GraphVAE in \cite{simonovsky2018graphvae} and our test of MolGAN \cite{de2018molgan} based on the codes available at https://github.com/nicola-decao/MolGAN. For MolGAN, we perfrom four different tests. The first one (no RL) applies MolGAN without reinforcement learning (RL). The next three (RL Valid, RL Unique and RL Novel) apply RL for validity, uniqueness and novelty, respectively. An RL step is done after each five epochs. For each setting we only report the best result among six choices of the parameter $\lambda$ (0.01, 0.05, 0.1, 0.25, 0.5, 0.75) and three choices of the dropout rate (0, 0.1, 0.25) used in \cite{de2018molgan}. Without RL, it takes {approximately 11 seconds} to train an epoch; with RL for validity or uniqueness, it takes 15-18s to train an epoch; with RL for novelty, it takes 93-122s to train an epoch.

\begin{table}[t]
\centering
\caption{Comparison of graph generation by GraphVAE, MolGAN and SCAT using the QM9 dataset. Values are reported in percentages according to 10k generated samples. Since the GraphVAE code is unavailable, results are copied from the paper and marked with parenthesis.}
\begin{tabular}{lccc}
Algorithm & Valid & Unique & Novel \\
\hline
GraphVAE &  (55.7) & (76.0) & (61.6) \\
GraphVAE (imp) & (56.2) & (42.0) & (75.8) \\
GraphVAE (no GM) &  (81.0) & (24.1) & (61.0) \\
\hline
MolGAN (no RL) & 90.4 & 31.1 & 97.8 \\
MolGAN (RL Valid) & 100.0 & 0.3 & 13.6 \\
MolGAN (RL Unique) & 99.2 & 37.1 & 64.5 \\
MolGAN (RL Novel) & 98.5 & 0.6 & 100.0 \\
\hline
{SCAT-SW} & 65.4 & 92.7 & 86.9 \\
{SCAT-DW} & 38.0 & 98.1 & 94.2 \\
{SCAT-SN} & 64.9 & 92.0 & 85.7 \\
{SCAT-DN} & 47.4 & 98.3 & 92.0
\end{tabular}
\label{tab:molcpr}
\end{table}

All the SCAT models achieve high scores in uniqueness and novelty. Note that SCAT-DW and SCAT-DN achieve slightly higher uniqueness and novelty than SCAT-SW and SCAT-SN, but they have notably lower score in validity. We remark that it is possible to achieve high validity with the scattering generative models. We were able to achieve 93.9/17.6/98.6 for valid/unique/novel scores if we train SCAT-SW using an MLP with a single hidden layer with dimension 5 for the vertices and 32 for the edges. However, a simpler model leads to severe mode collapse, as implied by the low uniqueness score. We observed that this simple model assigns ``carbon'' to a lot of vertices, and it has two effects: first, it is easier to construct valid molecules\footnote{A carbon vertex can have degree at most four while keeping the molecule valid. The largest possible degree is only three for a nitrogen vertex and two for an oxygen vertex.}. Second, it causes mode collapse as there is not much variety of molecular decompositions. 

\begin{figure*}[ht]
  \centering
  \begin{subfigure}[b]{.48\linewidth}
    \centering\includegraphics[width=\linewidth]{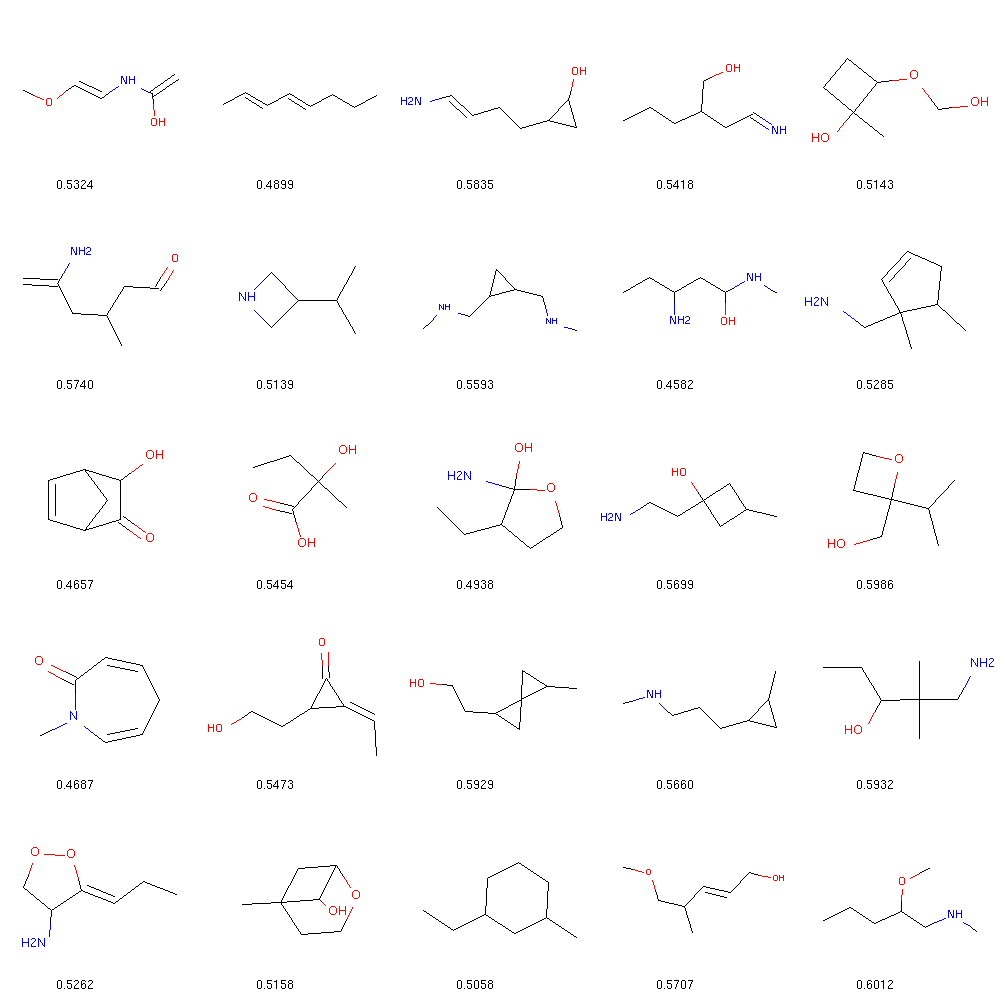}
    \caption{Samples from a generator with three hidden layers. 
    \label{fig:molscat}}
  \end{subfigure}%
  ~
  \begin{subfigure}[b]{.48\linewidth}
    \centering\includegraphics[width=\linewidth]{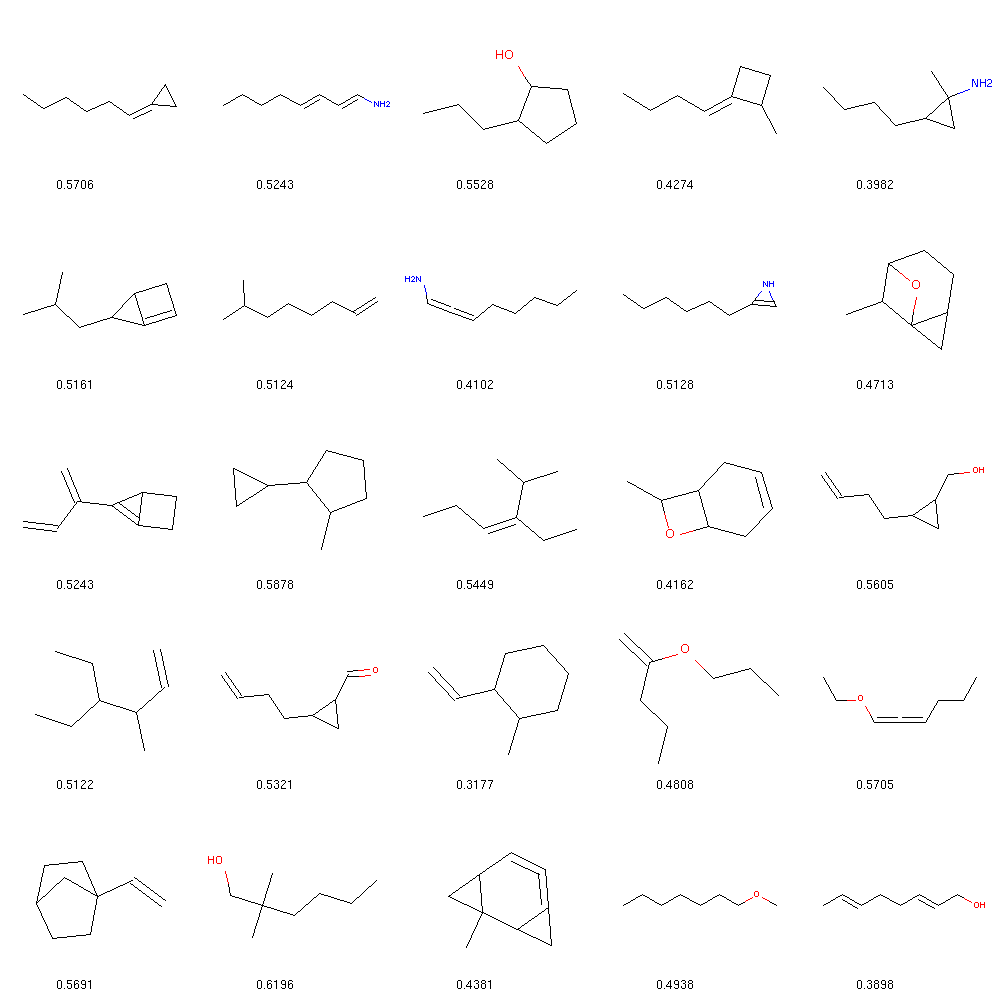}
    \caption{Samples from a generator with one hidden layer.  
  \label{fig:molsimple}}
  \end{subfigure}
  \caption{Samples of generated molecules via SCAT-SW. The number below each molecule is the Quantitative Estimate of Drug-likeness (QED) score \cite{bickerton2012quantifying}, which is automatically generated for the figures by RDKit. We are not trying to optimize it.}
  \label{fig:molcpr}
\end{figure*}

We show samples of molecules generated by SCAT-SW with a decoder with three hidden layers in Fig. \ref{fig:molscat} and samples generated by SCAT-SW with a decoder with one hidden layer in Fig. \ref{fig:molsimple}. It can be seen that many molecules in Fig. \ref{fig:molsimple} are composed only of carbon (and hydrogen); in contrast, there are more oxygen and nitrogen in the samples in Fig. \ref{fig:molscat}. We believe that in order to explore new design of drugs, it is more important to generate molecules with more variety. Therefore, the results from three-layer decoders are more meaningful.

As a comparison, the three measures reported for GraphVAE are moderate. MolGAN has good validity and novelty scores, but low uniqueness score, which indicates a mode collapse \cite{de2018molgan}. This mode collapse is even more severe when either validity or novelty is used for RL, although the model achieves perfect score for validity or novelty, respectively. We remark that MolGAN uses three hidden layers of the same dimensions and we thus believe that the mode collapse is not due to a low-complex structure of the generator.

\section{Conclusion}

We proposed the graph generative scattering network as a generative model for graphs and graph signals. The network applies a prescribed encoder which does not require training and is robust to signal perturbation and graph deformations. Numerical experiments show competitive results for the tasks of link prediction in citation data and molecule generation. Although scattering usually takes time, it is still more efficient to train scattering-based models due to the smaller number of parameters to update in training.
We believe the graph generative scattering network has the potential to be used in a wider range of applications on graphs.

We experimented with two possible choices for the scattering transform and two choices for Gaussianization. Overall, the scattering described in \cite{zou2018graph} achieves better performance than the scattering described in \cite{gama2018diffusion} for link prediction and molecule generation. We remark that for molecule generation the scattering of \cite{gama2018diffusion} achieves slightly better scores of uniqueness and novelty, however, its validity score is significantly worse. There is no significant difference in the results for the two methods of Gaussianization. However, the spherization-based methods (SCAT-SN and SCAT-DN) converge much faster for graph signal generation.

We used the Fashion-MNIST dataset as a sanity check for the graph signal generation, since we are unaware of a more convincing application for this specific task. In this application, we do not expect graph-based methods to compete with general methods, because graphs only retain partial spatial relationships. Indeed, the resolution of the generated images is not as good as that of the original images. Nevertheless, since the results in the similar discrete tasks of link prediction and molecule generation are competitive, we believe that SCAT also bears promise for graph signal generation when the signals are of low resolution.

\section*{Acknowledgment}
We thank Radu Balan, Aurobrata
Ghosh, and Maneesh Singh for discussions on the link prediction problem and Alex Gutierrez for helpful
comments on the paper.

\bibliography{mainRef}

\begin{thebibliography}{10}

\bibitem{zou2018graph}
D.~Zou and G.~Lerman, ``Graph convolutional neural networks via scattering,''
  {\em arXiv preprint arXiv:1804.00099}, 2018.

\bibitem{gama2018diffusion}
F.~Gama, A.~Ribeiro, and J.~Bruna, ``Diffusion scattering transforms on
  graphs,'' {\em arXiv preprint arXiv:1806.08829}, 2018.

\bibitem{henaff2015deep}
M.~Henaff, J.~Bruna, and Y.~LeCun, ``Deep convolutional networks on
  graph-structured data,'' {\em arXiv preprint arXiv:1506.05163}, 2015.

\bibitem{defferrard2016convolutional}
M.~Defferrard, X.~Bresson, and P.~Vandergheynst, ``Convolutional neural
  networks on graphs with fast localized spectral filtering,'' in {\em Advances
  in Neural Information Processing Systems}, pp.~3844--3852, 2016.

\bibitem{kipf2016semi}
T.~N. Kipf and M.~Welling, ``Semi-supervised classification with graph
  convolutional networks,'' in {\em International Conference on Learning
  Representations}, 2017.

\bibitem{chen2017supervised}
Z.~Chen, X.~Li, and J.~Bruna, ``Supervised community detection with
  hierarchical graph neural networks,'' {\em arXiv preprint arXiv:1705.08415},
  2017.

\bibitem{goodfellow2014generative}
I.~Goodfellow, J.~Pouget-Abadie, M.~Mirza, B.~Xu, D.~Warde-Farley, S.~Ozair,
  A.~Courville, and Y.~Bengio, ``Generative adversarial nets,'' in {\em
  Advances in neural information processing systems}, pp.~2672--2680, 2014.

\bibitem{kingma2013auto}
D.~P. Kingma and M.~Welling, ``Auto-encoding variational bayes,'' in {\em
  International Conference on Learning Representations}, 2014.

\bibitem{angles2018generative}
T.~Angles and S.~Mallat, ``Generative networks as inverse problems with
  scattering transforms,'' in {\em International Conference on Learning
  Representations}, 2018.

\bibitem{hammond2011wavelets}
D.~K. Hammond, P.~Vandergheynst, and R.~Gribonval, ``Wavelets on graphs via
  spectral graph theory,'' {\em Applied and Computational Harmonic Analysis},
  vol.~30, no.~2, pp.~129--150, 2011.

\bibitem{coifman2006diffusion}
R.~R. Coifman and M.~Maggioni, ``Diffusion wavelets,'' {\em Applied and
  Computational Harmonic Analysis}, vol.~21, no.~1, pp.~53--94, 2006.

\bibitem{sen2008collective}
P.~Sen, G.~Namata, M.~Bilgic, L.~Getoor, B.~Galligher, and T.~Eliassi-Rad,
  ``Collective classification in network data,'' {\em AI magazine}, vol.~29,
  no.~3, p.~93, 2008.

\bibitem{kipf2016variational}
T.~N. Kipf and M.~Welling, ``Variational graph auto-encoders,'' in {\em NIPS
  Workshop on Bayesian Deep Learning (NIPS-16 BDL)}, 2016.

\bibitem{xiao2017fashion}
H.~Xiao, K.~Rasul, and R.~Vollgraf, ``Fashion-{MNIST}: a novel image dataset
  for benchmarking machine learning algorithms,'' {\em arXiv preprint
  arXiv:1708.07747}, 2017.

\bibitem{ramakrishnan2014quantum}
R.~Ramakrishnan, P.~O. Dral, M.~Rupp, and O.~A. Von~Lilienfeld, ``Quantum
  chemistry structures and properties of 134 kilo molecules,'' {\em Scientific
  data}, vol.~1, p.~140022, 2014.

\bibitem{olivecrona2017molecular}
M.~Olivecrona, T.~Blaschke, O.~Engkvist, and H.~Chen, ``Molecular de-novo
  design through deep reinforcement learning,'' {\em Journal of
  cheminformatics}, vol.~9, no.~1, p.~48, 2017.

\bibitem{mallat2012group}
S.~Mallat, ``Group invariant scattering,'' {\em Communications on Pure and
  Applied Mathematics}, vol.~65, no.~10, pp.~1331--1398, 2012.

\bibitem{bruna2013invariant}
J.~Bruna and S.~Mallat, ``Invariant scattering convolution networks,'' {\em
  IEEE transactions on pattern analysis and machine intelligence}, vol.~35,
  no.~8, pp.~1872--1886, 2013.

\bibitem{bruna2013spectral}
J.~Bruna, W.~Zaremba, A.~Szlam, and Y.~LeCun, ``Spectral networks and locally
  connected networks on graphs,'' in {\em International Conference on Learning
  Representations}, 2014.

\bibitem{bojchevski2018netgan}
A.~Bojchevski, O.~Shchur, D.~Z{\"u}gner, and S.~G{\"u}nnemann, ``{N}et{GAN}:
  Generating graphs via random walks,'' in {\em Proceedings of the 35th
  International Conference on Machine Learning}, pp.~610--619, 2018.

\bibitem{wang2017graphgan}
H.~Wang, J.~Wang, J.~Wang, M.~Zhao, W.~Zhang, F.~Zhang, X.~Xie, and M.~Guo,
  ``{GraphGAN}: Graph representation learning with generative adversarial
  nets,'' in {\em AAAI Conference on Artificial Intelligence}, 2018.

\bibitem{de2018molgan}
N.~De~Cao and T.~Kipf, ``{MolGAN}: An implicit generative model for small
  molecular graphs,'' {\em arXiv preprint arXiv:1805.11973}, 2018.

\bibitem{simonovsky2018graphvae}
M.~Simonovsky and N.~Komodakis, ``Graph{VAE}: Towards generation of small
  graphs using variational autoencoders,'' {\em arXiv preprint
  arXiv:1802.03480}, 2018.

\bibitem{jin2018junction}
W.~Jin, R.~Barzilay, and T.~Jaakkola, ``Junction tree variational autoencoder
  for molecular graph generation,'' in {\em Proceedings of the 35th
  International Conference on Machine Learning}, pp.~2323--2332, 2018.

\bibitem{chen2001gaussianization}
S.~S. Chen and R.~A. Gopinath, ``Gaussianization,'' in {\em Advances in neural
  information processing systems}, pp.~423--429, 2001.

\bibitem{laparra2011iterative}
V.~Laparra, G.~Camps-Valls, and J.~Malo, ``Iterative gaussianization: from ica
  to random rotations,'' {\em IEEE transactions on neural networks}, vol.~22,
  no.~4, pp.~537--549, 2011.

\bibitem{bojanowski2018optimizing}
P.~Bojanowski, A.~Joulin, D.~Lopez-Pas, and A.~Szlam, ``Optimizing the latent
  space of generative networks,'' in {\em Proceedings of the 35th International
  Conference on Machine Learning}, pp.~600--609, 2018.

\bibitem{blum2017foundation}
A.~Blum, J.~Hopcroft, and R.~Kannan, {\em Foundations of Data Science}.
\newblock June 2017.

\bibitem{kingma2014adam}
D.~P. Kingma and J.~Ba, ``Adam: A method for stochastic optimization,'' {\em
  arXiv preprint arXiv:1412.6980}, 2014.

\bibitem{kusner2017grammar}
M.~J. Kusner, B.~Paige, and J.~M. Hern{\'a}ndez-Lobato, ``Grammar variational
  autoencoder,'' in {\em Proceedings of the 34th International Conference on
  Machine Learning}, pp.~1945--1954, 2017.

\bibitem{samanta2018designing}
B.~Samanta, A.~De, N.~Ganguly, and M.~Gomez-Rodriguez, ``Designing random graph
  models using variational autoencoders with applications to chemical design,''
  {\em arXiv preprint arXiv:1802.05283}, 2018.

\bibitem{bickerton2012quantifying}
G.~R. Bickerton, G.~V. Paolini, J.~Besnard, S.~Muresan, and A.~L. Hopkins,
  ``Quantifying the chemical beauty of drugs,'' {\em Nature chemistry}, vol.~4,
  no.~2, p.~90, 2012.

\end{thebibliography}
\bibliographystyle{ieeetr}

\end{document}